%% file: main.tex
\documentclass[sn-mathphys-num]{sn-jnl}

\usepackage[numbers]{natbib}

\usepackage{graphicx}%
\usepackage{amsmath,amssymb,amsfonts}%
\usepackage{amsthm}%
\usepackage{mathrsfs}%
\usepackage[title]{appendix}%
\usepackage{xcolor}%
\usepackage{textcomp}%
\usepackage{manyfoot}%
\usepackage{booktabs}%
\usepackage{algorithm}%
\usepackage{algorithmicx}%
\usepackage{algpseudocode}%
\usepackage{listings}%
\usepackage{multicol}

\usepackage[utf8]{inputenc}
\usepackage[T1]{fontenc}

\begin{document}

\title[Article Title]{Watch and Learn: Leveraging Expert Knowledge and Language for Surgical Video Understanding}


\author*[1,2]{\fnm{David} \sur{Gastager}}\email{davidgastager@gmail.com}

\author*[1]{\fnm{Ghazal} \sur{Ghazaei}}\email{ghazal.ghazaei@zeiss.com}

\author[2]{\fnm{Constantin} \sur{Patsch}}\email{constantin.patsch@tum.de}

\affil[1]{\orgdiv{Corporate Research and Technology}, \orgname{Carl Zeiss AG}, \orgaddress{\city{Munich}, \country{Germany}}}

\affil[2]{\orgname{Technical University of Munich}, \orgaddress{\city{Munich}, \country{Germany}}}

\abstract{
\textbf{Purpose:}
Automated surgical workflow analysis is a common yet challenging task with diverse applications in surgical education, research, and clinical decision-making. 
Although videos are commonly collected during surgical interventions, the lack of annotated datasets hinders the development of accurate and comprehensive workflow analysis solutions.
We introduce a novel approach for addressing the sparsity and heterogeneity of annotated training data inspired by the human learning procedure of \textit{watching experts and understanding their explanations}.

\textbf{Methods:}
Our method leverages a video-language model trained on alignment, denoising, and generative tasks to learn short-term spatio-temporal and multimodal representations. 
A task-specific temporal model is then used to capture relationships across entire videos.
To achieve comprehensive video-language understanding in the surgical domain, we introduce a data collection and filtering strategy to construct a large-scale pretraining dataset from educational YouTube videos.
We then utilize parameter-efficient fine-tuning by projecting downstream task annotations from publicly available surgical datasets into the language domain. 

\textbf{Results:}
Extensive experiments in two surgical domains demonstrate the effectiveness of our approach, with performance improvements of up to 7\% in phase segmentation tasks, 8\% in zero-shot phase segmentation, and comparable capabilities to fully-supervised models in few-shot settings.
Harnessing our model’s capabilities for long-range temporal localization and text generation, we present the first comprehensive solution for dense video captioning (DVC) of surgical videos, addressing this task despite the absence of existing DVC datasets in the surgical domain.

\textbf{Conclusion:}
We introduce a novel approach to surgical workflow understanding that leverages video-language pretraining, large-scale video pretraining, and optimized fine-tuning. Our method improves performance over state-of-the-art techniques and enables new downstream tasks  for surgical video understanding.
}

\keywords{Multi-model Video Understanding, Vision-Language Models, Surgical Workflow Recognition, Dense Video Captioning}
\maketitle

%
%
%
%
%
\section{Introduction}
\textbf{Automated surgical workflow analysis} has the potential to greatly improve patient care by enhancing safety, reducing time and costs, and improving training. 
Surgical videos, widely captured during procedures and used as educational materials, are key to developing such solutions. 
However, progress is limited by the scarcity of annotated datasets, as the annotation process often requires \textbf{expert knowledge}. 
The diversity of procedures and clinical settings further complicates creating datasets that are generalizable across domains, leading to task- and domain-specific models.

Recent advancements in computer vision achieved great success with pretraining or self-training approaches on large-scale datasets, leading to the \textbf{emergence of generalist models}.
In natural language processing, models like BERT\cite{Bert} and GPT\cite{GPT} use masked language modeling or next-token prediction to train large transformer architectures. 
This was further followed by the introduction of \textbf{vision language models (VLMs)} where multimodal representation learning unlocked new possibilities, especially with text or visual prompting. 
For instance, CLIP\cite{Clip} aligns language and image spaces using a contrastive approach, while other multimodal methods have extended to video understanding using self-supervised techniques such as contrastive span training\cite{Reserve} and multimodal group captioning\cite{Valor}. 
Models like \cite{Reserve,Valor} leverage large-scale web videos with captions or ASR-transcribed speech for training. 
More recently, efforts have been made in creating \textbf{surgical foundation models}. EndoViT\cite{Endo700k}, for instance, trains a vision transformer on a dataset with over 700.000 endoscopy images. 
SurgVLP\cite{SurgVLP} and OphCLIP\cite{ophclip} train CLIP-style video-text models on large datasets of intervention videos with text transcribed using ASR.
In this work, we adopt \textbf{natural language as an intermediate representation} to bring datasets to the same annotation space, leveraging their unique features across downstream tasks.
We employ a \textbf{two-stage architecture} where stage~1 serves as a \textbf{video-language model that builds short-range dependencies} between densely sampled video frames, while stage~2 consists of a task-specific \textbf{temporal model to build long-range dependencies} across the video.
Our video-language model leverages a modified VALOR\cite{Valor} as a foundation multimodal architecture for video-language learning which we pretrain on over \textbf{2800 educational cataract surgery videos with expert commentary} from YouTube\cite{youtube} extracted using ASR.
We employ a \textbf{multimodal pretraining strategy} with generative, denoising, and alignment tasks, showing that training on large-scale heterogeneous data with cross-modality learning promotes a robust, procedure-specific representation generalizable across various acquisition setups and clinics.
\\
Generalist models can provide useful general representations, but they may face challenges in handling domain-specific knowledge, while traditional fine-tuning procedures in low-data regime setups can suffer from catastrophic forgetting. 
\textbf{Low-rank adaptation (LoRA)} \cite{Lora} is an alternative fine-tuning technique that was first introduced to incorporate new information into large language models via low-rank decomposition matrices\cite{Lora}. 
We exploit LoRA to effectively integrate domain-specific knowledge, preserving general representations while adapting to fine-tuning data.
Processing full-length surgical videos poses challenges due to their long duration and the need for fine-grained analysis of brief phases, like those lasting only one second in cataract surgery. 
This requires a model capable of handling both pixel-level detail and video-level procedure understanding. 
Many video feature learning architectures\cite{Valor,Reserve,SurgVLP,ophclip} are limited to short-range analysis, as they operate on short input clips. 
Long-range reasoning typically involves heavy subsampling or freezing parts of the model.

Frameworks for \textbf{dense video captioning (DVC)} on lengthy videos, such as Vid2Seq\cite{vid2seq} or PDVC \cite{PDVC}, utilize a frozen pretrained feature extractor that encapsulates frame features in a single token and additionally heavily subsample the input video. 
This inevitably leads to the loss of short-range temporal information and patch-level granularity.
To address this challenge, we adopt a common approach in surgical workflow recognition\cite{glmstcn,Asformer} that concatenates a short-term spatio-temporal feature extractor with a long-term temporal model\cite{Tcn,Asformer} to effectively incorporate both types of information into the model.
For the task of DVC, which requires more fine-grained information, we take advantage of our predicted phases via our two-stage approach and feed all video frames inside one predicted phase back to our short-term video-language model.
Given the absence of publicly available datasets for DVC in any surgical domain, we only evaluate our solution qualitatively.

\textit{Our contributions} can be summarized as follows: \\
\textbf{Data Collection}: Propose a strategy for collecting and filtering a large-scale video-language dataset of cataract surgeries with expert commentary enabling learning a generic multimodal representation for cataract procedures.\\
\textbf{Pipeline}: Propose a two-stage architecture to model short- and long-range video-language relations. 
Stage~1 leverages a multimodal pretraining strategy to generate a global video-language representation, while stage~2 harnesses vision-guided language tokens for the downstream task. 
Propose a training pipeline that effectively integrates multiple datasets using low-rank adaptation.\\
\textbf{Downstream Performance}: Evaluate the method across various tasks, demonstrating strong generalizability, achieving state-of-the-art (SOTA) results in phase segmentation for cataract and cholecystectomy surgeries, and SOTA performance in zero-shot phase segmentation in cataract surgery.\\
\textbf{Surgical Dense Video Captioning} Propose a solution for the previously unexplored downstream task of dense video captioning in the surgical field, despite the lack of a DVC dataset in this domain.  \\

%
%
%
%
%
%
%
%
%
\section{Methodology}
\subsection{Vision-Language Modeling} \label{sec:vlm}
\begin{figure*}[t!]
    \centering
    \includegraphics[width=\linewidth]{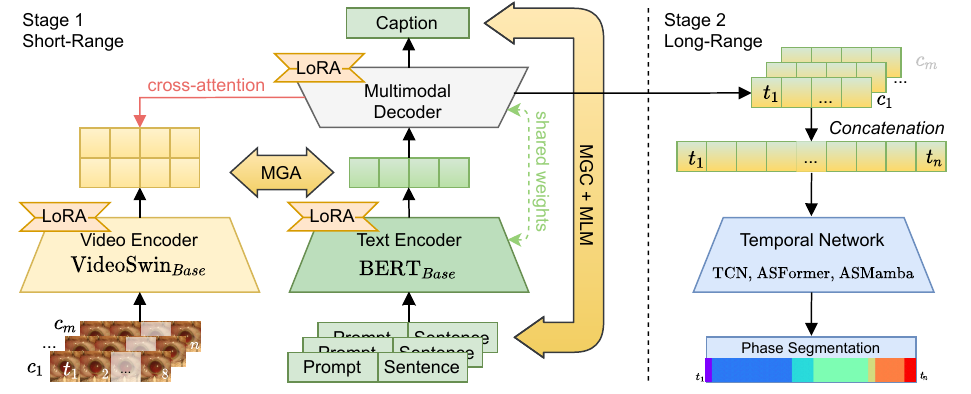}
    \caption{Overview of proposed two-stage architecture: \textbf{Stage~1)} A modified VALOR\cite{Valor} video-language model processes clips of $N$ frames while maintaining short-range dependencies. It is pretrained on our large-scale video-language dataset and fine-tuned on target datasets using LoRA \cite{Lora}.
    \textbf{Stage~2)} A temporal model \cite{Tcn,Asformer,videomambasuite} captures global interactions over a whole video. $t_i$ refers to frame timestamp and $c_i$ to clip number.}
    \label{fig:2stage_architecture}
\end{figure*}
We employ VALOR$_B$\cite{Valor}, a video-language model consisting of three key components: VideoSwin$_B$\cite{VideoSwin} as a video-encoder, BERT$_B$\cite{Bert} as a text encoder, and a BERT-style multimodal decoder which shares its weights with the text encoder, but is equipped with additional cross-attention layers. The multimodal decoder also includes causal masking and autoregressive text generation.
We drop the audio encoder proposed in the original VALOR architecture.
The left side of Fig.~\ref{fig:2stage_architecture} shows a conceptual overview. 
We follow the VALOR's video-language pretraining strategy and pretrain stage~1 using three tasks.
\\
\textbf{1) Multimodal Group Alignment (MGA)}
$N_v$ frames are uniformly sampled from the input clip and fed to the video-encoder to create spatio-temporal video tokens $\mathbf{V} \in \mathbb{R}^{N_v \times S_v \times C_v}$, where $S_v$ corresponds to the number of spatial tokens and $C_v$ to the video encoders hidden dimension. 
The concatenated task prompt and input text are passed through the text encoder to create text tokens $\mathbf{T} \in \mathbb{R}^{N_t \times C_t}$, where $N_t$ is the number of text tokens and $C_t$ the hidden dimension. 
The MGA-loss, a batch-wise bidirectional contrastive loss for aligning video and text embeddings, is defined as:
\begin{equation} \label{eq:mga_loss} \scriptstyle
    L_{MGA}(\mathbf{T},\textbf{V}) = - \frac{1}{2} \sum\limits_{i=1}^{B} \log \frac{\exp ( s(T_i, V_i)/\tau)}{\sum_{j=1}^B \exp(s(T_i, V_j)/\tau} 
     - \frac{1}{2} \sum\limits_{i=1}^{B} \log \frac{\exp ( s(T_i, V_i)/\tau)}{\sum_{j=1}^B \exp(s(T_j, V_i)/\tau} ,
\end{equation} 
with similarity function $s(.,.)$ and a learnable temperature parameter $\tau$.
The similarity computation is done in a fine-grained manner between all text and video tokens:
\begin{equation} \label{eq:mga_sim} \scriptstyle
    s(T,V) = \frac{1}{2} \sum\limits_{i=1}^{N_t} f_{t, \theta}(e_t)\max\limits_{j=1}^N(e_t^i)^\top e_v^j 
    + \frac{1}{2} \sum\limits_{i=1}^{N_v} f_{v, \theta}(e_v) \max\limits_{j=1}^N(e_t^i)^\top e_v^j ,
\end{equation}
where $f_{i, \theta}$ is a learnable weighing of the tokens for vision $(e_v)$ and text $(e_t)$ and $N_i$ is the number of tokens of each modality. The similarity function is the summation of bidirectional scores between each modality and is computed by maximizing the similarity along one matrix dimension \cite{Valor}.
\\
\textbf{2) Multimodal Group Captioning (MGC)} 
masks a large portion of the text input before the text encoder processes it.
The masked tokens are then passed to a multimodal decoder, which leverages cross-attention to incorporate video-encoder information and autoregressively generate a caption. 
The MGC-loss, a cross-entropy loss with causality constraints, assesses the quality of the generated caption:
$L_{MGC}(\mathbf{T}, \mathbf{V}) = - \mathbb{E}_{(T,V) \in B} \log P(T_m|T_{<m}, V)$
where $B$, $T_m$, $T_{<m}$ denote the training batch, the masked tokens and the previously given tokens. 
\\
\textbf{3) Masked Language Modeling (MLM)} masks only a small percentage of the text tokens and uses cross-entropy loss:
\( L_{MLM}(\mathbf{t}, \hat{\mathbf{t}}) = -\sum_{i \in I_m} t_i \cdot \log \hat{t}_i \) (\( I_m \) as indices, \({\mathbf{t}}\) and \(\hat{\mathbf{t}}\) as true and predicted token probabilities).

This leads to the final VALOR loss as $L_{VALOR} = \frac{1}{3}(L_{MGA} + L_{MGC} + L_{CE})$
\subsection{Downstream Tasks}
\subsubsection{Phase Segmentation}
\textbf{Two-stage approach}
Fig.~\ref{fig:2stage_architecture} illustrates our proposed solution. In \textbf{stage~1},
a video is first split into $m$ non-overlapping clips $\{c_1,...,c_m\}$ of equal length and processed per clip by the stage~1 video-language model to extract vision-aware language tokens and their \textbf{short-range} dependencies. 
These tokens are then concatenated and fed to the \textbf{second stage} to model \textbf{long-range} dependencies over the whole video.
To match the input dimensions of the respective stage~2 model, a convolutional pooling layer and a 1x1 convolution with ReLU activations are added.
\newline
\textbf{Temporal Models} \label{sec:temp_models}
We experiment with three different temporal models in stage~2:
\begin{itemize}
    \item \textbf{MS-TCN++}\cite{Tcn} (TCN): featuring dilated temporal convolutions and multiple stages of refinement
    \item \textbf{ASFormer}\cite{Asformer} (ASF): an encoder-decoder transformer model using dilated temporal convolutions and cross-attention layers to refine predictions.
    \item \textbf{ASMamba}\cite{videomambasuite} (ASM): An ASFormer variant that uses Mamba State-Space Layers\cite{mamba} instead of self-attention in the transformer encoder.
\end{itemize}
\textbf{Zero-Shot Phase Prediction} 
Training stage~1 with MGA aligns video and text embeddings for CLIP-style zero-shot classification \cite{Clip}.
Embedding the target classes in prototype sentences and feeding them through the text encoder enables cosine similarity calculation between video and text embeddings. 
The class with the highest similarity is selected as the prediction, and frame-by-frame results are concatenated for full video segmentation.

\subsubsection{Dense Video Captioning} \label{sec:dvc}
For dense video captioning we combine the segmentation and captioning capabilities of our method.
We first feed an input video to the phase segmentation model to predict all event boundaries, followed by grouping the frames of each predicted phase into consecutive chunks of 10 seconds. 
These are then fed to the video-language model to generate a caption sentence. 
The resulting sentences are collected with their respective start and end timestamps to create the final video transcript.
We do not produce captions for predicted \textit{Idle} phases.
\subsection{Data Collection \& Preparation} 
Despite the absence of publicly available video-language datasets for surgical videos, we train a cataract surgery video-language model using a two-pillar data strategy:
First, we collect a large video-language dataset of cataract surgeries with expert commentary from publicly available sources.
Second, we use existing video datasets with phase and tool annotations, projecting them into the language domain. 

\subsubsection{Pretraining Dataset Collection Procedure}
We build a pretraining dataset for cataract surgery from publicly accessible YouTube videos with expert commentary. 
We use the YouTube API \cite{youtubeapi} to search for videos related to cataract surgery (e.g. \textit{cataract surgery}, \textit{IOL placement}, \textit{capsulorhexis}) and ophthalmologist-run channels. Videos are filtered based on metadata to ensure relevant content, using a predefined list of \textit{in-words} (e.g. \textit{cataract}, \textit{phacoemulsification}) and excluding \textit{out-words} (e.g. \textit{podcast}, \textit{Q\&A}, \textit{Tech Review}). 
The audio commentary of the selected videos is transcribed using the Whisper$_L$\cite{Whisper} model. We correct common medical terminology errors using a lookup table. 
Since these videos are made for a general audience, we manually identify and aim to remove non-surgical segments via the addressing the following:
\textbf{1) people presenting on camera} A face detection algorithm is used to classify frames with faces as non-valid.
\textbf{2) static presentation slides} A frame is deemed non-valid if fewer than half its pixels differ from the previous frame, accounting for typical camera noise in natural videos. In presentation slides, minimal pixel changes often occur due to compression artifacts.
\textbf{3) text overlays} We detect bounding boxes around textual elements and apply a stochastic greedy search to find an optimal crop that excludes all text. 
Frames cropped smaller than 224x224 are classified as non-valid. 
We apply a three-second temporal median filter to all validity classifications to finalize the valid video regions.  
Since our stage~1 model is trained on short video-language pairs, we divide videos into clips based on speech punctuation, discarding clips shorter than 2 seconds. 
\\
From here on, we will refer to our collected dataset as the \textbf{YT-dataset (YT)}.
\begin{table}[t!]
    \centering
    \fontsize{9}{10}\fontfamily{ptm}\selectfont
    \begin{tabular*}{\linewidth}{@{\extracolsep{\fill}}| l |  r || l | r |}
        \hline
        \multicolumn{2}{|c||}{\textbf{YT Dataset Statistics}} & \# Clips & 149.939\\
        \# Videos & 2.933 & Avg. Clip Duration & 6.5s\\
        Avg. Video Duration & 7min 33s & Total Clip Duration & 268hours 10min 4s\\
        Total Video Duration & 369hours 31min 43s & \# Words & 2.247.750\\
        Filtered Video Duration & 333hours, 58min, 25s & Avg. \# Words per Clip & 15 \\
        \hline
    \end{tabular*} 
    \caption{Statistics of the collected video-language dataset of cataract surgeries (YT dataset).}
    \label{tbl:YT_Dataset}
\end{table}

\subsection{Dataset Integration \& Low Rank Adaptation:}
To bring available annotated datasets to the language domain, we embed their labels into template sentences tailored to each dataset task. 
For the tool and phase detection tasks in a cataract surgery dataset, we insert \textit{\texttt{TOOL}} and \textit{\texttt{PHASE}} labels into the sentence \textit{"The surgeon is using a \texttt{TOOL} during the \texttt{PHASE} phase of cataract surgery."}. This allows training our stage~1 model on all annotated data sets.
To effectively pretrain our video-language model across multiple datasets and prevent catastrophic forgetting, we use \textbf{low-rank adaptation}. LoRA layers are integrated into all attention layers of the stage~1 model alongside the query and value weight matrices.

%
%
%
%
%
%
%
%
\section{Implementation Details}
\subsection{Datasets}
\textbf{Cataract Surgery}
The CATARACTS\cite{Cataracts} dataset (CAT) comprises 50 cataract surgery videos annotated for phase segmentation and temporal tool localization, with a 25/5/20 split for training, validation, and testing, following \cite{Cataracts, FelixCataracts}. It includes 22 tool and 19 phase classes, including \textit{Idle} class usually present between phases.
Cataract-101\cite{Cataract101} (C101) contains 101 cataract surgery videos annotated with 11 phase labels. Unlike CAT, the \textit{Idle} phase appears only at the start of the videos. Following \cite{C101SOTA}, we split the dataset into 73 videos for training/validation and 28 for testing.
\newline
\textbf{Cholecystectomy Surgery}
Cholec80 \cite{cholec80} is an endoscopic video dataset containing 80 videos of cholecystectomy surgery with annotations for phase segmentation with seven phases and tool presence for seven tools with a 40/40 split for training and validation following \cite{sftmn}.
The CholecT50 \cite{cholecT50} dataset comprise 45 videos of Cholec80 and 4 new videos, all with additional text annotations. It includes \texttt{<TOOL,VERB,TARGET>} triplets for 100 different actions classes which we embed into the prototype sentence \textit{"The surgeon is using a \texttt{TOOL} to \texttt{VERB} the \texttt{TARGET}."}.
We exclude videos from the Cholec80 test and validation sets from the CholecT50 train set to avoid data leakage.

\subsection{Training Procedure}
\textbf{Stage~1} \label{stage1_models}
In the cataract surgery domain, we first pre-train the video-language model on our collected YT dataset for 20 epochs using a batch size of 384. The learning rate anneals from $10^{-4}$ to $10^{-9}$ after one epoch warmup. 
We mask 60\% of text tokens for MGC and 10\% for MLM, following \cite{Valor}. 
We freeze the model's parameters and activate its LoRA layers to train on the language-projected CAT dataset. We use a batch size of 24. The learning rate anneals from $10^{-2}$ to $10^{-4}$. 
As there are no large-scale video-language dataset available in the cholecystectomy domain, we only utilize CholecT50\cite{cholecT50} for pre-training stage~1 using the same hyperparameters as for training on YT. 
We perform experiments using the following variants of our stage~1 model, all initialized with the pretrained VALOR$_B$ \cite{Valor} weights:
\begin{itemize}
    \item \texttt{V-YT}: Pre-training on collected YT dataset
    \item \texttt{V-CAT}: Pre-training on language-projected CAT dataset
    \item \texttt{V-YT-FT-CAT}: Pre-training on YT, fine-tuning on language-projected CAT with \textbf{full weight} updates
    \item \texttt{V-YT-LoRA-CAT}: Pre-training on YT, fine-tuning on language-projected CAT using \textbf{low-rank adaptation}
    \item \texttt{V-CT50}: Pre-training on language-projected CholecT50 dataset
\end{itemize}
\textbf{Stage~2}
Extracted features from the frozen stage~1 model are used to train a temporal model (see section \ref{sec:temp_models}) on each of the three target datasets. 
We use a batch size of 1 and a learning rate annealing from $10^{-3}$ to $10^{-7}$.

All hyperparameters can be found in the supplementary material.
\subsection{Evaluation Metrics}
All evaluations are conducted at 1 fps to ensure consistency with existing literature. Following previous studies \cite{sftmn, glmstcn}, we assess performance using frame-wise accuracy and per-phase metrics, including precision, recall, and Jaccard index, averaged at the \textbf{video level}. We also report the F1 score. 
To evaluate temporal consistency, we use additional metrics \cite{Tcn}, including the Edit score (EDIT), and Overlap F1 at different thresholds $\tau$. 
To assess the prediction accuracy of our model on a \textbf{frame-level} across the entire dataset, we also compute the Accuracy averaged over all frames (Acc$_{\text{micro}}$).
For Cholec80, we evaluate the performance of our model in the \textbf{unrelaxed boundary setting}. A comparison against SOTA methods in the relaxed boundary setting can be found in the supplementary materials.
%
%
%
%
%
%
%
%
\section{Results \& Discussion}
\begin{table*}[b!]
    \centering
    \resizebox{\linewidth}{!}{
    \begin{tabular}{ l c | c c c c c | c |c c c c c }

    \multicolumn{13}{c}{\textbf{(A) CATARACTS Dataset}} \\
    \hline
    \multicolumn{2}{c|}{Method} & \multicolumn{5}{c|}{Video-level metrics} & & \multicolumn{5}{c}{Temporal metrics}  \\
    Stage~1 & Stage~2 & Accuracy & Precision & Recall & Jaccard & F1 & Acc$_{\text{micro}}$ & EDIT & O-F1@10 & @25 & @50 & Avg O-F1 \\
    \hline
        DINO\cite{Dino}* & TCN & 77.2$\pm$12.4 & 78.3$\pm$12.0 & 75.5$\pm$12.0 & 60.5$\pm$15.4 & 72.0$\pm$11.9 & 72.3 & 70.9 & 72.7 & 69.7 & 61.1 & 67.8\\
        \multicolumn{2}{c|}{Holm et. al. \cite{FelixCataracts}} & 75.2 & - & - & - & 68.6 & - & - & - & - & - & - \\
        \multicolumn{2}{c|}{Sangria \cite{koksal2024sangria}} & 83.4 & - & - & - & 78.3 & - & - & - & - & - & - \\
        ResNet34 & SR-Mamba\cite{srmamba} & - & - & - & - & 85.4 & - & - & - & - & - & - \\
        \hline
        \multicolumn{2}{c}{\textbf{Ours}} & \multicolumn{11}{c}{}\\
        \hline
        V-CAT & TCN & 87.7$\pm$8.8 & 89.0$\pm$10.4 & 86.4$\pm$12.3 & 77.8$\pm$14.3 & 85.5$\pm$8.0 & 83.9 & 83.4 & 84.3 & 81.9 & 78.0 & 81.4\\
        V-YT & TCN & 85.7$\pm$10.4 & 85.9$\pm$9.6 & 85.8$\pm$9.4 & 74.8$\pm$12.4 & 82.9$\pm$9.1 & 80.1 & 81.2 & 82.3 & 81.6 & 74.6 & 79.5\\
        V-YT-FT-CAT & TCN  & 88.2$\pm$7.1 & 88.1$\pm$8.6 & 87.8$\pm$10.3 & 77.9$\pm$12.5 & 85.0$\pm$9,4 & 84.9 & 83.0 & \textbf{86.1} & 84.8 & 79.7 & 80.5\\ 
        V-YT-LoRA-CAT & TCN & {88.9$\pm$7.1} & 88.2$\pm$8.9 & 88.3$\pm$8.6 & 78.8$\pm$10.9 & 86.1$\pm$7.4& {85.6} & \textbf{83.9} & 86.0 & \textbf{84.9} & \textbf{80.8} & \textbf{83.9}\\
        V-YT-LoRA-CAT & ASF & 88.9$\pm$7.2 & {89.3$\pm$9.4} & {89.4$\pm$8.3} & {80.7$\pm$10.2} & {87.6$\pm$6.5} & {85.6} & 81.7 & 84.6 & 83.6 & 79.3 & 82.5\\
        \textbf{V-YT-LoRA-CAT} & ASM & \textbf{90.3$\pm$6.6} & \textbf{90.6$\pm$7.9} & \textbf{89.6$\pm$8.0} & \textbf{81.9$\pm$10.3} & \textbf{88.2$\pm$5.9} & \textbf{87.7} & 82.1 & 85.0 & 84.6 & 80.6 & 83.4\\
    \hline

    \multicolumn{13}{c}{}\\
    \multicolumn{13}{c}{\textbf{(B) Cataract-101 Dataset}} \\
        \hline
        \multicolumn{2}{c|}{STANet\cite{glmstcn}} & 95.3 & 93.4 & 93.5 & 87.9 & - & - & - & - & - & - & - \\
        GL-MSTCN\cite{glmstcn} & TCN & 96.5 & 94.9 & 95.2 & 90.8 & - & - & - & - & - & - & -\\
        \hline
        \multicolumn{2}{c}{\textbf{Ours}} & \multicolumn{11}{c}{}\\
        \hline
        V-CAT & TCN & 93.1$\pm$5.5 & 92.3$\pm$4.9 & 90.9$\pm$5.5 & 84.6$\pm$8.3 & 90.3$\pm$6.5 & 92.5 & 89.4 & 92.9 & 92.1 & 87.3 & 90.8 \\
        V-YT & TCN & 94.4$\pm$3.8 & 93.3$\pm$4.8 & 92.0$\pm$5.7 & 86.0$\pm$8.8 & 90.9$\pm$7.0 & 94.4 & 91.8 & 94.6 & 94.0 & 89.9 & 92.8 \\
        V-YT-LoRA-CAT & TCN & 96.1$\pm$2.4 & 94.6$\pm$3.0 & 94.2$\pm$3.6 & 89.2$\pm$5.7 & 93.3$\pm$4.4 & 95.6 &\textbf{ 94.1} &\textbf{96.3} & \textbf{96.0} & \textbf{93.5} & \textbf{95.3} \\
        V-YT-LoRA-CAT & ASF & {97.0$\pm$2.5} & \textbf{95.9$\pm$3.0} & {96.3$\pm$2.6}& {92.5$\pm$4.9} & {95.7$\pm$3.3} & {96.4} & 92.4 & 94.9 & 94.9 & 92.4 &  94.1\\
        V-YT-LoRA-CAT & ASM & \textbf{97.3$\pm$1.8} & 95.8$\pm$2.9 & \textbf{96.7$\pm$1.9}&\textbf{92.7$\pm$4.4} & \textbf{95.7$\pm$3.1} & \textbf{97.0} & 91.8 & 95.4 & 95.1 & \textbf{93.5} &  94.7\\
    \hline

    \multicolumn{13}{c}{}\\
    \multicolumn{13}{c}{\textbf{(C) Cholec80 Dataset} (unrelaxed boundary metrics)} \\
        \hline
        \multicolumn{2}{c|}{LoViT\cite{lovit}} & 91.5$\pm$6.1 & 83.1 & 86.5 & 74.2 & - & - & - & - & - & - & - \\
        \multicolumn{2}{c|}{SKiT\cite{skit}} & 92.5$\pm$5.1 & 84.6 & 88.5 & 76.7 & - & - & - & - & - & - & - \\
        \multicolumn{2}{c|}{SurgFormer\cite{surgformer}} & 92.4$\pm$6.4 & 87.9$\pm$6.9 & 89.3$\pm$7.8 & 79.9$\pm$10.2 & - & - & - & - & - & - & - \\
        \hline
        \multicolumn{2}{c}{\textbf{Ours}} & \multicolumn{11}{c}{}\\
        \hline
        \textbf{V-CT50} & TCN & 94.0$\pm$6.5 & 87.8$\pm$7.6 & 91.0$\pm$7.5 & 81.6$\pm$9.9 & 88.0$\pm$7.9 & 92.9 & \textbf{88.8} & \textbf{91.6} & \textbf{91.3} & 85.5 & \textbf{89.5}\\
        \textbf{V-CT50} & ASF &\textbf{95.1}$\pm$\textbf{3.2} & \textbf{88.2}$\pm$\textbf{7.3} & \textbf{91.4}$\pm$\textbf{7.0} & \textbf{82.5}$\pm$\textbf{9.3} & \textbf{88.5$\pm$7.4} & \textbf{94.6} & 87.0 & 90.6 & 90.6 & \textbf{86.0} & 89.1\\
        V-CT50 & ASM & 94.5$\pm$4.9 & 87.5$\pm$8.4 & 90.8$\pm$7.6 & 82.0$\pm$10.3 & 87.8$\pm$8.4 & \textbf{94.6} & 74.2 & 78.0 & 77.7 & 72.8 & 76.2\\
    \hline
    
    \multicolumn{13}{c}{} \\
    \multicolumn{13}{c}{\textbf{(D) Cataract-101 Zero-Shot} (Stage~1 only)} \\
        \hline
        \multicolumn{2}{c|}{OphCLIP\cite{ophclip}} & 39.3 & - & - & - & 34.2 & - & - & - & - & - & - \\
        \hline
        V-CAT & - & 33.0$\pm$9.3 & 27.0$\pm$6.5 & 25.6$\pm$6.3 & 12.9$\pm$4.1 & 19.2$\pm$5.7 & 33.0 & 2.6 & 1.5 & 0.8 & 0.3 & 0.9\\
        V-YT & - & 19.8$\pm$4.7 & 26.9$\pm$6.4 & 25.9$\pm$5.0 & 13.5$\pm$4.3 & 18.6$\pm$4.8 & 19.7 & \textbf{3.1} & 1.9 & 1.2 & \textbf{0.5} & 1.2\\
        V-YT-FT-CAT & - & 33.0$\pm$7.0 & 29.5$\pm$6.1 & 39.5$\pm$5.1 & 18.2$\pm$4.2 & 27.2$\pm$5.4 & 32.9 & 2.1 & 2.0 & 1.3 & 0.5 & 1.3 \\
        V-YT-LoRA-CAT & - & \textbf{47.4$\pm$7.6} & \textbf{48.5$\pm$5.2}&\textbf{44.9$\pm$6.0} & \textbf{28.2$\pm$5.0} & \textbf{40.1$\pm$5.7} & \textbf{47.9} & 2.5 & \textbf{2.6} &\textbf{1.5}& 0.4 & \textbf{1.5} \\
        \hline
    \end{tabular}
    }
    \caption{Performance comparison of our proposed method on the CATARACTS, Cataract-101 and Cholec80 datasets: 
    $\pm$ indicates the standard deviation of the respective metrics between all videos. 
    *The DINO-TCN implementation uses a DINO\cite{Dino}-trained ViT$_{Base}$ with patch size 16.
    }
    \label{tab:benchmark_cataract}
\end{table*}
\subsection{Phase Segmentation}
\textbf{CATARACTS}
Table~\ref{tab:benchmark_cataract} (A) demonstrates the superior performance of our proposed method compared to other approaches on the CAT dataset. 
Compared to our baseline implementation of a DINO\cite{Dino}-TCN combination, our worst model V-YT already outperforms it by over 8\%, highlighting the effectiveness of our stage~1 architecture. 
Better accuracy of V-YT than Sangria\cite{koksal2024sangria} ($\sim 4.5\%$) highlights the efficacy of video-language representation learning and pretraining on our YT-dataset.
Our results show that training on a large-scale dataset of YouTube videos (V-YT) yields competitive performance to training directly on the target dataset (V-CAT).
This highlights the diversity of the learned features attributed to the large variety of collected videos and our suggested video-language representation learning approach. 
The impact of large-scale pretraining can be further observed in fine-tuning. 
LoRA fine-tuning (V-YT-LoRA-CAT) achieves better results than full fine-tuning (V-YT-FT-CAT), which we attribute to LoRA's ability to effectively guide weight updates while preserving the model's original weights. 
That is, all LoRA models exhibit an emergence effect, capturing large-scale information from the YT dataset and integrating the underlying structure of CAT, setting a new frame-level baseline of $\sim 90\%$. 
It is also clearly highlighted in zero-shot experiments (Table~\ref{tab:benchmark_cataract}).
Our best setup (V-YT-LoRA-CAT-ASM) outperforms the SOTA SR-Mamba\cite{srmamba} by $\sim 3\%$ in F1-score, and Sangria\cite{koksal2024sangria}
by a significant margin of 7\% and 10\% in accuracy and F1 score, respectively.
\\
\textbf{Cataract-101}
\begin{figure}[b!]
    \centering
    \includegraphics[width=0.9\linewidth]{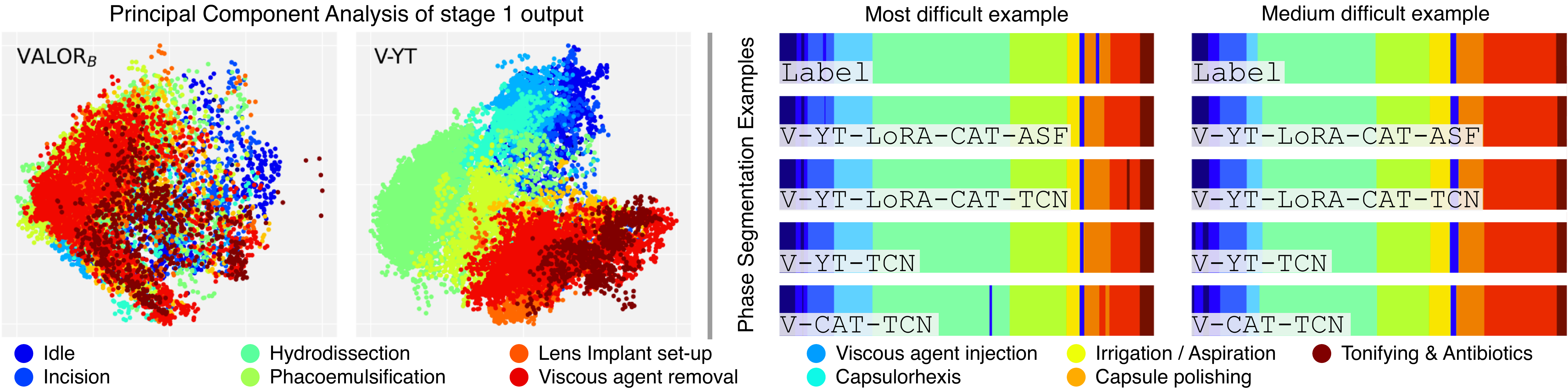}
    \caption{\textit{Left:} The first two PCA components of the stage 1 output before (left) and after (right) training on our YT-dataset.
    \textit{Right:} Example phase predictions of different variants of our method. \textit{Bottom:} Label legend. All examples on C101.}
    \label{fig:pca}
\end{figure} 
Fig.~\ref{fig:pca} (left) shows the first two PCA components on C101 before and after YT pretraining, highlighting the model's zero-shot clustering ability and its effectiveness in capturing key cataract surgery stages. 
As highlighted in Table~\ref{tab:benchmark_cataract} (B), V-CAT underperforms compared to V-YT, which can be accounted for the substantial domain gap between CAT and C101, as well as the greater diversity of features learned by V-YT and the lack of extended expert commentary.
Each model iteration indicates the contribution of added components to the architecture, also visible in Fig.~\ref{fig:pca} (right).
This is particularly noteworthy for C101 for which the stage~1 was never specifically fine-tuned. 
This further emphasizes the impact of video-language understanding together with large-scale pretraining setting a robust foundation for new domains.
V-YT-LoRA-CAT leverages YT and CAT pretraining to boost the performance up to $\sim 3\%$ across all metrics. 
The previous state-of-the-art GL-MSTCN\cite{glmstcn} uses surgery-specific branches for tool and pupil processing.
Despite the lack of such branches and our stage~1 not being trained on C101, our V-YT-LoRA-CAT-ASF and -ASM models outperform GL-MSTCN across all metrics. 
An additional ablation study on the generalization abilities can be found in the supplementary materials.

\textbf{Cholec80} In Table~\ref{tab:benchmark_cataract} (C), our model variants consistently outperform the SOTA, underscoring the remarkable potential of video-language modeling. Notably, this performance is achieved despite being pretrained solely on the limited language information provided by CholecT50, and without the benefit of large-scale pretraining. Among the variants, the ASF model exhibits superior video-level metrics, while the TCN model excels in temporal performance, likely due to the temporal smoothing effects introduced by its convolutional kernels, which is particularly beneficial for the long phases of the Cholec80 dataset. When compared to the prior SOTA model SurgFormer\cite{surgformer}, which employs a sophisticated hierarchical temporal attention mechanism, our V-CT50-ASF demonstrates significant improvements, setting a new baseline of $95.1\%$ video-level accuracy. We believe that the alignment of video and language features imposes a structured organization on the representations, enabling the stage~2 model to categorize phases more effectively by leveraging language tokens as explicit and interpretable contextual cues. Examples are in the supplementary material.
\\
\textbf{Zero-Shot Phase Prediction} \label{sec:zero_shot}
Table~\ref{tab:benchmark_cataract} (D) highlights a zero-shot evaluation of different model variants on C101.
V-YT, V-CAT and V-YT-FT-CAT perform inadequately on zero-shot phase prediction for the unseen dataset of C101 showcasing the distinct characteristics of each dataset.
V-YT-LoRA-CAT exhibits notably superior performance compared to the other models, emphasizing the emergence effect even more evidently. Compared to its fully fine-tuned equivalent (V-YT-FT-CAT), it is better equipped to utilize and merge information from both datasets while retaining previously learned knowledge from initial training which facilitates better zero-shot generalizability. 
Our best model V-YT-LoRA-CAT outperforms the previous SOTA OphCLIP\cite{ophclip} by a large margin of ~8\% in accuracy and ~6\% in F1 score, despite OphCLIP being trained on a substantially larger dataset of over 7,500 hours. This further illustrates the efficacy of our data-centric approach and emphasizes the value of training on annotated datasets by projecting them into the language domain.
Due to the absence of a stage~2 component in these experiments, the temporal consistency of the results is lower than that of the two-stage variants.
Still, the V-YT-LoRA-CAT captures the main trend of surgical phases of C101 while never being trained on that nor having distinct phase definitions other than that of CAT. 
Qualitative examples of zero-shot phase predictions can be found in the supplementary material.
\subsection{Dense Video Captioning}
\begin{figure*}[ht!]
    \centering
    \includegraphics[width=0.9\linewidth]{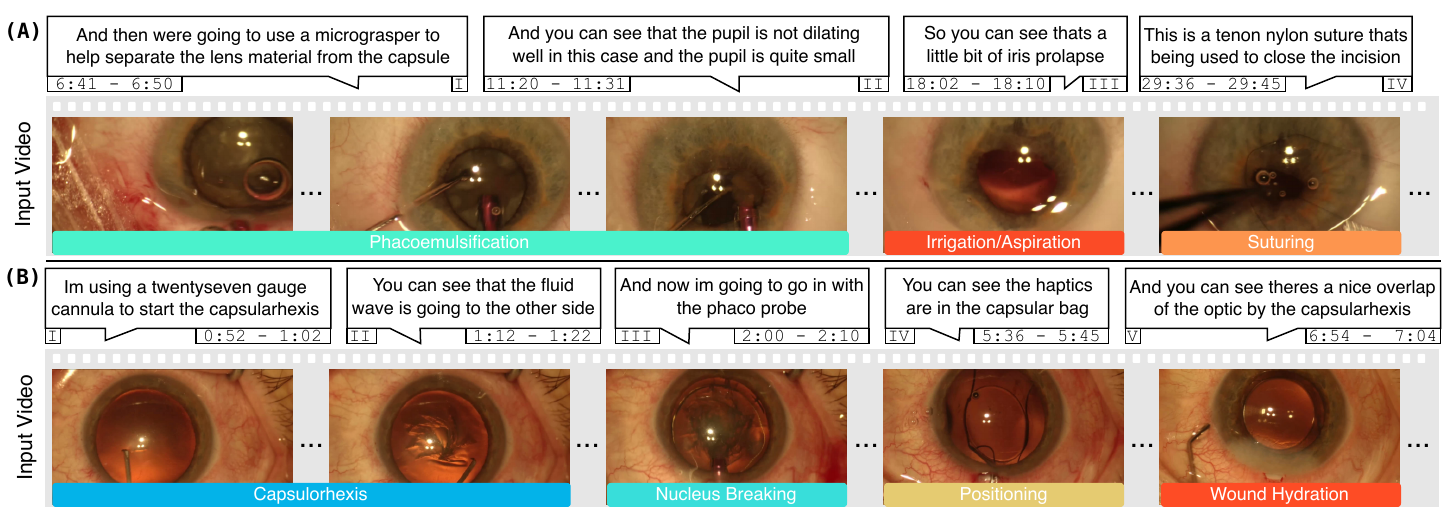}
    \caption{
    Example captions and phase predictions on two videos of the CATARACTS test set. \texttt{(A)} is from video~2, which includes a surgery complication (iris prolapse). \texttt{(B)} represents parts of video~8. The fully captioned videos are available in the supplementary materials.
    }
    \label{fig:caps_qual}
\end{figure*}

We exploit V-YT for dense video captioning on the CAT dataset, taking advantage of predicted phases from V-YT-LoRA-CAT-TCN to generate temporally consistent and diverse captions, that maintain coherence and relevance to the underlying video content (Fig.~\ref{fig:caps_qual}).
As there are no DVC datasets available for any surgical domain, we evaluate the captions qualitatively. 
The proposed model demonstrates strong capabilities in generating detailed captions for surgical procedures. It accurately identifies and describes delicate features, such as the fluid wave in the capsularhexis (Fig.\ref{fig:caps_qual}\texttt{(B)II}), and provides descriptions of surgical tools and their uses (Fig.\ref{fig:caps_qual}\texttt{(A)I+IV}). Furthermore, the model effectively identifies rare complications, such as iris prolapse (Fig.\ref{fig:caps_qual}\texttt{(A)III}), and provides context by describing preceding conditions, such as poor pupil dilation (Fig.\ref{fig:caps_qual}\texttt{(A)II}), highlighting the holistic understanding of our captioning model. It can also comprehensively describe all 
steps of a cataract surgery, along with the tools and actions involved. Fig.~\ref{fig:caps_qual}(B) describes the capsularhexis, nucleus breaking, artificial lens positioning and wound hydration phases of cataract surgery with impressive details.
However, we find three main limitations of the model. First, it occasionally hallucinates details, such as lengths or sizes, without a proper frame of reference (e.g., Fig.\ref{fig:caps_qual}\texttt{(B)I}), likely due to dataset biases where tools are frequently present with specific measurements. Second, it sometimes omits action descriptions, such as wound hydration in Fig.\ref{fig:caps_qual}\texttt{(B)V}, focusing instead on qualitative analysis like the position of the lens. Thirdly, it sometimes repeats descriptions inside a single caption. 
These limitations can be addressed with more extensive data collection and verification.
Our results indicate that large-scale video-language training not only improves generalizability and performance but also opens up new possibilities for surgical video analysis. A more detailed analysis can be found in the supplementary material.

%
%
%
%
%
%
\section{Conclusion}
We demonstrate a novel two-stage solution for surgical workflow understanding leveraging video-language pretraining, large-scale video pretraining, and optimized fine-tuning. 
We highlight the versatility and effectiveness of our approach by improved performance on three public datasets representing two distinct surgical domains and two downstream tasks. We present SOTA performance in phase segmentation and introduce dense video captioning for cataract videos.
Our findings underscore the importance of large-scale video-language comprehension in capturing the comprehensive semantics of surgical videos, providing a diverse range of opportunities for automated surgical workflow analysis.

\input{main.bbl}

\clearpage

\appendix
\setcounter{table}{2}
\setcounter{figure}{3}
\renewcommand{\thepage}{\roman{page}}
\setcounter{page}{1}
\section{\texttt{Supplementary Material for} \\
Watch and Learn: Leveraging Expert Knowledge and Language for Surgical Video Understanding}
\subsection{Hyperparameters}
During training, input videos from the cataract domain are sub-sampled at 5 fps, while the cholecystectomy experiments use 1fps subsampling, following \cite{cholec80}.
\subsubsection*{Stage~1}
For training stage~1, we linearly subsample eight frames from each input clip.
We use the prompt \textit{"Describe the video with natural language"} and mask 60\% of the text tokens for MGC, following \cite{Valor}. 
Initially we tried masking 100\%, but this turned out to be too difficult for the model to properly learn to create captions. 
For MLM we mask 10\% of text tokens.
For MGA, we use the task prompt \textit{"Project the inputs into common space"}.

The model is trained for 20 epochs using the AdamW optimizer with a cosine annealing learning rate scheduler with one epoch warmup. 
The maximum and minimum values for the learning rate are $10^{-4}$ and $10^{-9}$. 
We use a batch size of 6 on each GPU and accumulate gradients for 32 steps, resulting in a total batch size of 384.
To stabilize training, the model's gradients are clipped if they exceed a value of 5.
The same settings are used for LoRA training, but the minimum and maximum learning rates are set to $10^{-2}$ and $10^{-4}$, while the effective batch size is 24.
The LoRA rank $r$ is set to $8$.

The resulting model has a total of 239 million parameters, with 87M for the video encoder, 137M for text encoder and multimodal decoder, and 15M for the contrastive heads and dimensionality adjustment layers. 
The additional LoRA layers for finetuning only consist of 14M parameters.
\subsubsection*{Stage 2}
The MS-TCN++ uses a hidden dimension of 64, a maximum kernel size of 25 and has 3 refinement stages. 
It is trained using cross entropy loss and has 24M parameters. 
The ASFormer model also uses a hidden dimension of 64, but uses equally weighted cross entropy and dice loss. 
It consists of 9 encoder layers and 3 decoder layers, and needs 1.4M parameters. 
The ASMamba variant is similar to ASFormer, but uses Mamba-DBM\cite{videomambasuite} layers instead of self-attention layers in the encoder.
The temporal models are trained for 150 epochs using the AdamW optimizer and we use the same learning rate scheduling strategy as for stage~1.
The minimum and maximum learning rates are $10^{-7}$ and $10^{-3}$. 
As we train on full videos, a batch size of 1 is used.
\clearpage
\subsubsection*{Training Time}
All models were trained using four NVIDIA Tesla V100 32GB GPUs. Table \ref{tab:train_times} shows the training times of our different model variants.
\begin{table}[h!]
    \centering
    \begin{tabular}{l|c|c}
        Model & Training Dataset & approximate Training Time \\
        \hline
        V-YT & YT & 2 days 8 hours\\
        V-CAT & CAT & 2 hours 30 min\\
        V-YT-FT-CAT (Finetuning only) & CAT & 1 hour 15 min\\
        V-YT-LoRA-CAT (LoRA training only) & CAT & 45 min\\
        V-CT50 & CholecT50 & 4 hours 30 min \\
        TCN & CAT/C101/Cholec & 30 min \\
        ASF & CAT/C101/Cholec & 4 hours \\
        ASM & CAT/C101/Cholec & 4 hours \\
    \end{tabular}
    \caption{Training Times of our different architectures.}
    \label{tab:train_times}
\end{table}

\subsection{Ablation Study on Generalization}
To assess the generalization ability of our method, we designed an ablation study in which we used a frozen version of our best stage~1 model (V-YT-LoRA-CAT) together with TCN model in stage~2 trained on incrementally increasing portions of the training data. For model details see section 3.2 of the paper.
The TCN weights were initialized with the model weights trained on the CAT. 

Table~\ref{tab:subset_training} shows the quantitative results. 
The 0\% experiment utilized a TCN trained on CAT and employing a fixed label mapping from CAT to C101 labels, based on similar class definitions. 
The 1\% model performed worse than the 0\% model in most metrics, likely due to overfitting on the very small subset. 
However, the 10\% model already learned the most relevant features, leading to high 80\% range metrics. Adding additional videos improved the model's performance trained on the 50\%, 90\%, and 100\% subsets.  
Fig.\ref{fig:c101_examples} \texttt{(B)} confirms this outcome quantitatively, where the model with only 10\% training data performs competitively with the model with 100\% of labels on a difficult video of C101 (indicating small border inconsistencies only).
This highlights the generalizability of our data-rich video language modeling in stage~1, where minimal data from the target dataset is necessary to achieve solid performance. 
This has significant implications for the medical community, where access to data and annotation can be a major challenge. 

\begin{table*}[h!]
    \centering
    \resizebox{\linewidth}{!}{
    \begin{tabular}{ l | c c c c c | c |c c c c c }
    \multicolumn{12}{c}{\textbf{Cataract-101 Subset Training}} \\
    \hline
    Annotation Subset & \multicolumn{5}{c|}{Video-level metrics} & & \multicolumn{5}{c}{Temporal metrics} \\
     (Percentage) & Accuracy & Precision & Recall & Jaccard & F1 & Accuracy$_{\text{micro}}$ & EDIT & F1@10 & @25 & @50 & Avg O-F1 \\
    \hline
        0\% & 58.9$\pm$13.7 & 72.8$\pm$8.2 & 48.9$\pm$12.0 & 37.7$\pm$12.2 & 51.8$\pm$9.9 & 63.4 & 34.1 & 43.9 & 38.9 & 23.1 & 35.3 \\
        1\% & 41.9$\pm$13.6 & 50.0$\pm$13.8 & 36.1$\pm$11.4 & 23.4$\pm$8.0 & 55.3$\pm$9.8 & 38.1 & 59.7 & 43.2 & 36.1 & 20.1 & 33.1\\
        10\% & 87.3$\pm$6.8 & 83.8$\pm$6.9 & 82.5$\pm$6.3 & 70.4$\pm$8.6 & 79.8$\pm$6.7 & 86.6 & 87.0 & 89.1 & 85.3 & 74.9 & 83.1\\
        50\% &94.4$\pm$4.5 & 92.6$\pm$4.7 & 92.3$\pm$5.0 & 85.9$\pm$7.9 & 91.2$\pm$5.9 & 93.7 & 93.4 & 95.0 & 94.8 & 91.2 & 93.6\\
        90\% & 95.3$\pm$4.8 & 93.6$\pm$5.0 & 93.9$\pm$5.5 & 88.0$\pm$9.1 & 92.2$\pm$7.2 & 94.7 & 91.4 & 93.6 & 92.8 & 89.6 & 92.0\\
        100\% & \textbf{96.1$\pm$2.4} &\textbf{ 94.6$\pm$3.0} & \textbf{94.2$\pm$3.6} & \textbf{89.2$\pm$5.7} & \textbf{93.3$\pm$4.4} &\textbf{ 95.6} &\textbf{ 94.1} &\textbf{96.3} & \textbf{96.0} & \textbf{93.5} & \textbf{95.2}\\
    \hline
\end{tabular}
}
\caption{\textbf{Ablation}: Cataract-101 test results of two-stage solution (V-YT-LoRA-CAT-TCN) trained with different amounts of C-101 training data.}
\label{tab:subset_training} 
\end{table*}

\subsection{Additional Results on Cholec80}
Table \ref{tab:cholec80_relaxed} provides a detailed comparison of our proposed method with state-of-the-art (SOTA) approaches on the Cholec80 dataset under the \textbf{relaxed boundary evaluation} setting. Our method establishes a new benchmark, achieving an accuracy of $96.3\%$, which surpasses the previous SOTA method, SF-TMN-ASF \cite{sftmn}, by $+0.9\%$. Notably, our approach also demonstrates significant improvements in precision ($+3.2\%$) and Jaccard index ($+2.7\%$), while maintaining comparable performance in recall ($-0.1\%$).

In comparison to other recent methods, such as SurgFormer \cite{surgformer}, SKiT \cite{skit}, SR-Mamba \cite{srmamba}, and LoViT \cite{lovit}, our approach achieves a notable margin of improvement, with approximately $+3\%$ in accuracy, $+4\%$ in precision, and $+4\%$ in Jaccard index. These results underline the effectiveness of our method, also setting a new standard for phase recognition on the Cholec80 dataset in the relaxed boundary evaluation.
\begin{table*}[h!]
    \centering
    \resizebox{0.7\linewidth}{!}{
    \begin{tabular}{ l c | c c c c }
    \multicolumn{6}{c}{\textbf{Cholec80 Dataset} (relaxed boundary metrics)} \\
    \hline
    \multicolumn{2}{c|}{Method} & \multicolumn{4}{c}{Video-level metrics}\\
    Stage~1 & Stage~2 & Accuracy & Precision & Recall & Jaccard\\
    \hline
        \multicolumn{2}{c|}{LoViT\cite{lovit}} & 92.4$\pm$6.3 & 89.9$\pm$6.1 & 90.6$\pm$4.4 & 81.2$\pm$9.1\\
        \multicolumn{2}{c|}{SR-Mamba\cite{srmamba}} & 92.6$\pm$8.6 & 90.3$\pm$5.2 & 90.6$\pm$7.2 & 81.5$\pm$8.6\\
        \multicolumn{2}{c|}{SKiT\cite{skit}} & 93.4$\pm$5.2 & 90.9 & 91.8 & 82.6 \\
        \multicolumn{2}{c|}{SurgFormer\cite{surgformer}} & 93.4$\pm$5.2 & 91.0$\pm$4.7 & 92.1$\pm$5.8 & 84.1$\pm$8.0\\
        SF-TMN\cite{sftmn} & TCN & 93.6$\pm$4.7 & 91.2$\pm$4.5 & 90.7$\pm$6.5 & 83.0$\pm$6.3\\
        SF-TMN\cite{sftmn} & ASF & 95.4$\pm$4.0 & 92.4$\pm$5.3 & \textbf{93.4}$\pm$\textbf{4.4} & 86.1$\pm$6.6\\ 
        \hline
        \multicolumn{2}{c}{\textbf{Ours}} & \multicolumn{4}{c}{}\\
        \hline
        V-CT50 & TCN & 95.4$\pm$6.5 & 95.1$\pm$3.0 & 92.8$\pm$4.0 & 87.8$\pm$6.0 \\
        \textbf{V-CT50} & \textbf{ASF} & \textbf{96.3}$\pm$\textbf{3.1} & 94.8$\pm$5.3 & 92.5$\pm$4.9 & 87.7$\pm$8.0 \\
        \textbf{V-CT50} & \textbf{ASM} & 96.0$\pm$4.8 & \textbf{95.6}$\pm$\textbf{4.1} & 93.3$\pm$4.7 & \textbf{88.8}$\pm$\textbf{6.9}\\
    \hline
    \end{tabular}
    }
    \caption{Performance comparison of our proposed method on the Cholec80 datasets using the relaxed boundary metrics, following: \href{https://github.com/YuemingJin/TMRNet/blob/main/code/eval/result/matlab-eval/Evaluate.m}{\texttt{github.com/YuemingJin/TMRNet}} \\
    $\pm$ indicates the standard deviation of the respective metrics between all videos.
    }
    \label{tab:cholec80_relaxed}
\end{table*}

\subsection{Phase Segmentation Examples}
\subsubsection{Cataract-101}
\begin{figure}[h!]
    \centering
    \includegraphics[width=0.8\linewidth]{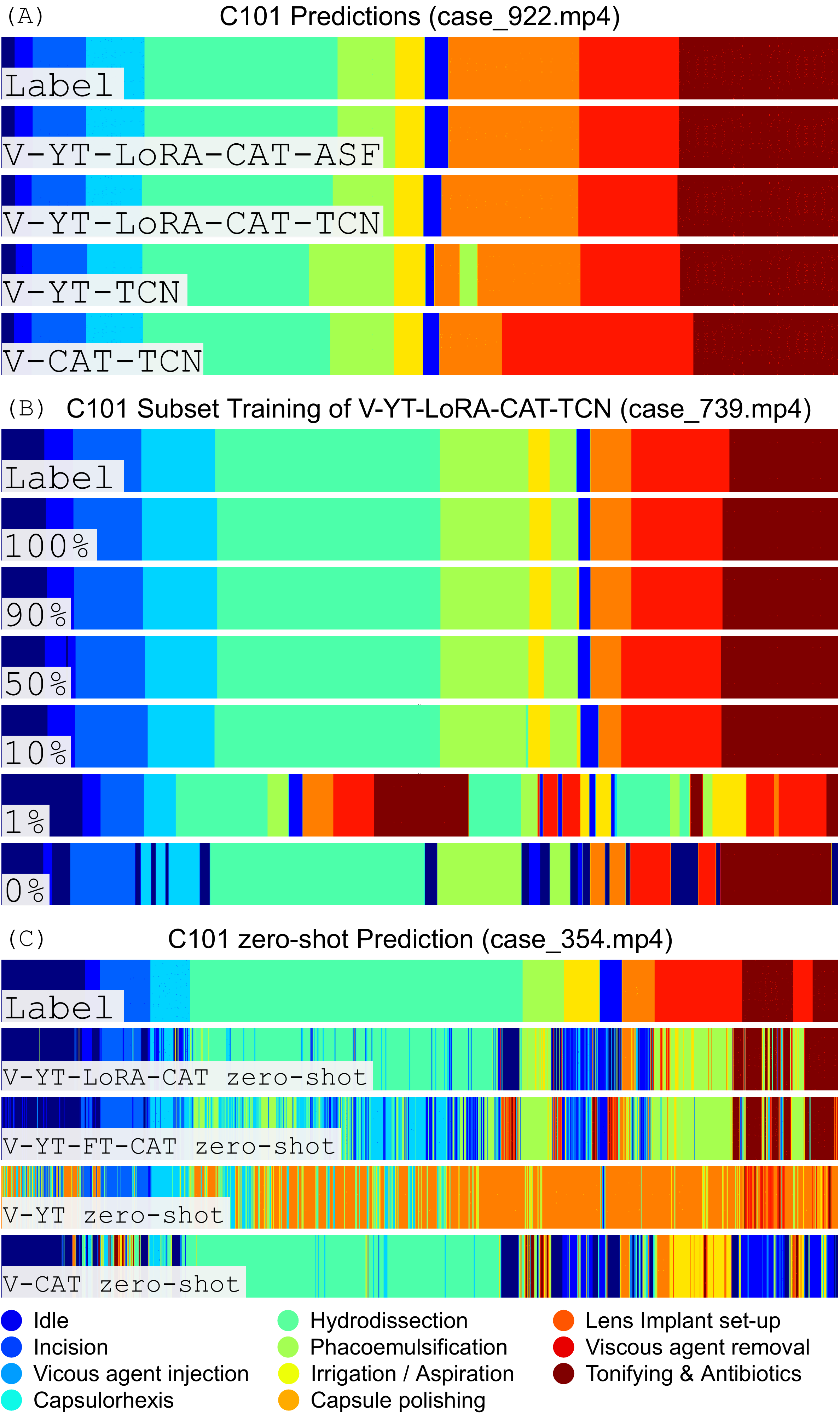}
    \caption{\texttt{(A)} Predictions of our model variants on a difficult video ($\sim8$~min duration, expert surgeon) from the C101 test set. 
    \texttt{(B)} Predictions of the \texttt{V-YT-LoRA-CAT-TCN} model trained with different subsets of the C101 training set on a difficult video ($15.5$~min duration, novice surgeon).
    \texttt{(C)} Zero-Shot predictions of different stage~1 models on a medium difficulty video of C101 ($\sim7.5$~min, novice surgeon). The zero-shot predictions lack temporal consistency, as they did not include a stage~2 model. The impact of LoRA is still clearly visible capturing the general trend of the phase segmentation.}
    \label{fig:c101_examples}
\end{figure}

Figure \ref{fig:c101_examples} presents phase segmentation predictions from various versions of our method on samples from the C-101 test set. 
Figure \ref{fig:c101_examples} \texttt{(A)} displays the predictions of our model variants on a challenging video (approximately 8 minutes long, performed by an expert surgeon). 
The models that use V-YT-LoRA-CAT in stage1 predict the segmented phases nearly perfectly, while the variants using V-YT and V-CAT in stage1 both struggle with the \textit{lens implant set-up} and \textit{viscous agent removal} phases.

Figure \ref{fig:c101_examples} \texttt{(C)} further demonstrates the strength of low-rank adaptation in effectively merging information from multiple datasets into a single model. 
The fully fine-tuned variant V-YT-FT-CAT produces a very noisy prediction, particularly during the \textit{hydrodissection} phase. 
In contrast, the LoRA variant, V-YT-LoRA-CAT, leverages information from the CAT dataset more efficiently, resulting in a much more confident prediction during the \textit{hydrodissection} phase.

\newpage

\clearpage
\subsubsection{Cholec80}
\begin{figure}[h!]
    \centering
    \includegraphics[width=\linewidth]{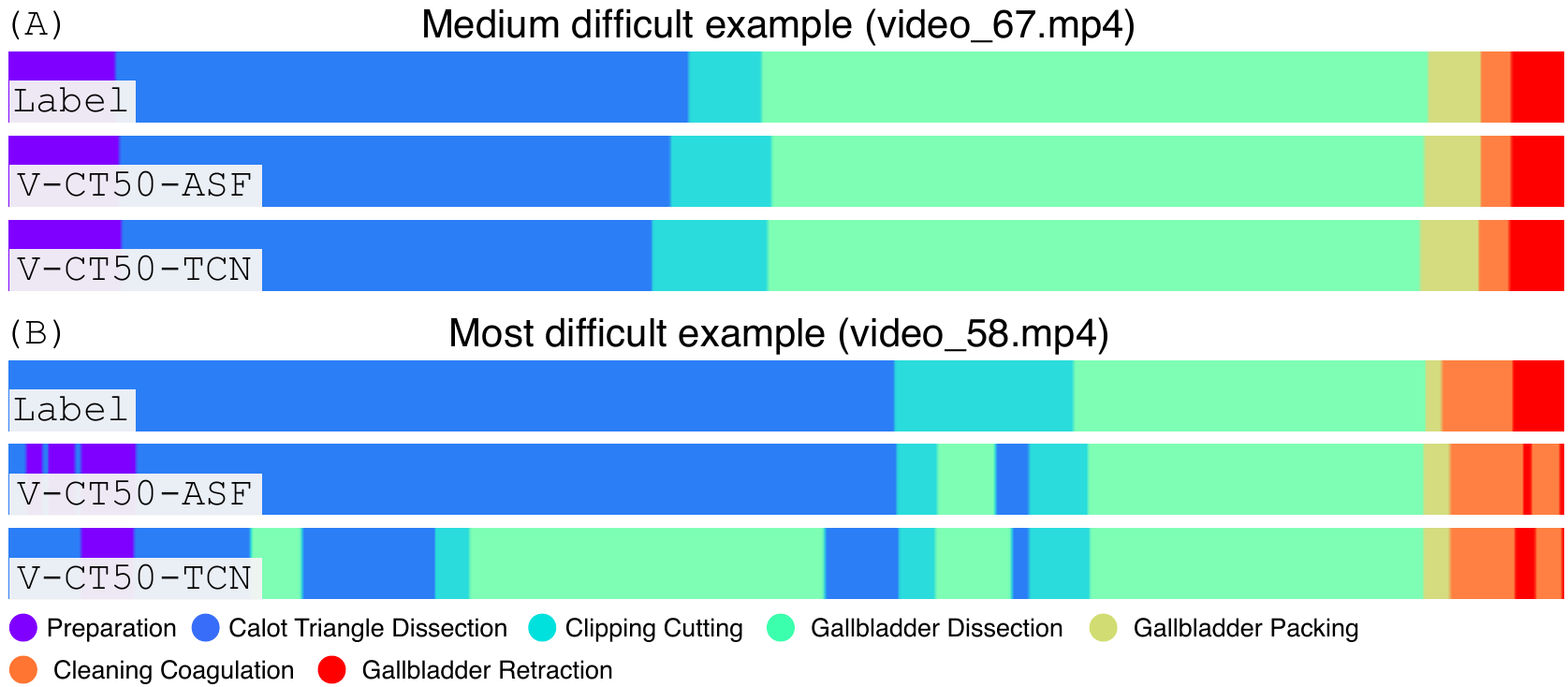}
    \caption{Example predictions of V-CT50-ASF and V-CT50-TCN on a medium difficult (\texttt{(A)}: video\_67.mp4) and the most difficult (\texttt{(B)}: video\_58.mp4) examples from the Cholec80 test set. Difficulty is relative and was measured by the achieved F1 scores by the models.}
    \label{fig:cholec_example}
\end{figure}

Figure~\ref{fig:cholec_example} shows phase segmentation predictions from two variants on two examples from the Cholec80 test set. On the most difficult example (\texttt{(B)}: video\_58.mp4), both variants struggle to identify the middle and end phases correctly. However, V-CT50-ASF predicts the Calot Triangle Dissection more accurately. On the medium difficult example (\texttt{(A)} video\_67.mp4), the models predictions are almost perfect with only a slight border inaccuracy of V-CT50-ASF. On both videos the TCNs predictions are slightly worse, showcasing the superior performance of ASF. This is also in line with the results described in table 1 of the main paper. Difficulty is relative, therefore the most difficult example was chosen based on the F1-score performance of our models.

\subsection{Dense Video Captioning Examples}
\begin{figure}[h!]
    \centering
    \includegraphics[width=0.98\linewidth]{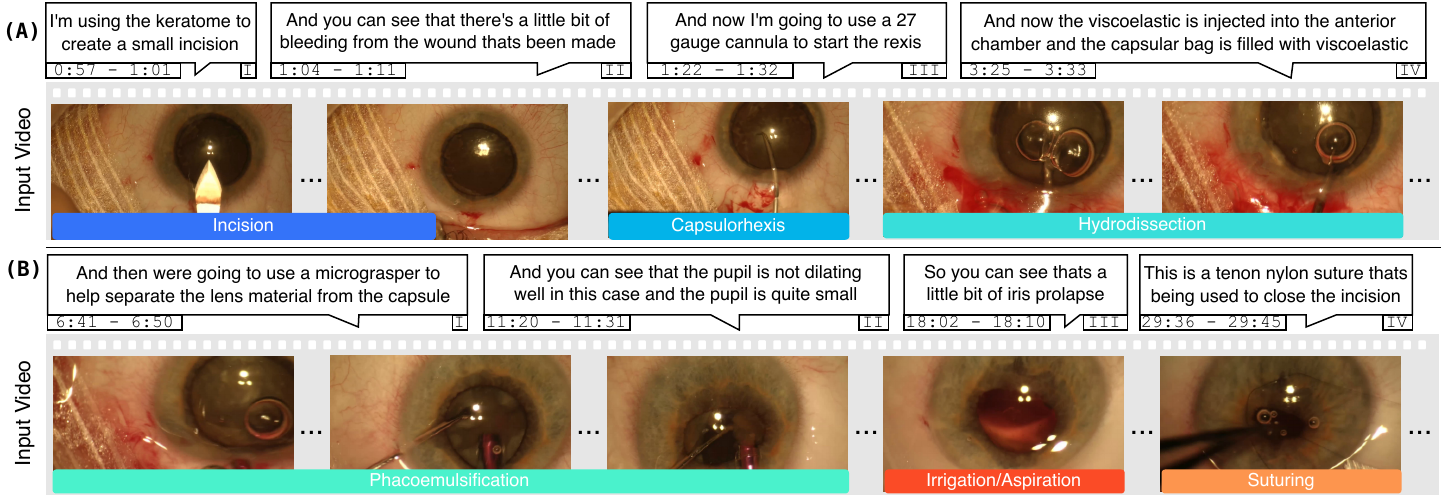}
    \caption{
    Example captions and phase predictions on video \textit{test\_02.mp4} of the CATARACTS test set. The fully captioned video is available in the supplementary materials.
    }
    \label{fig:caps_qual_02}
\end{figure}

\begin{figure}[h!]
    \centering
    \includegraphics[width=0.98\linewidth]{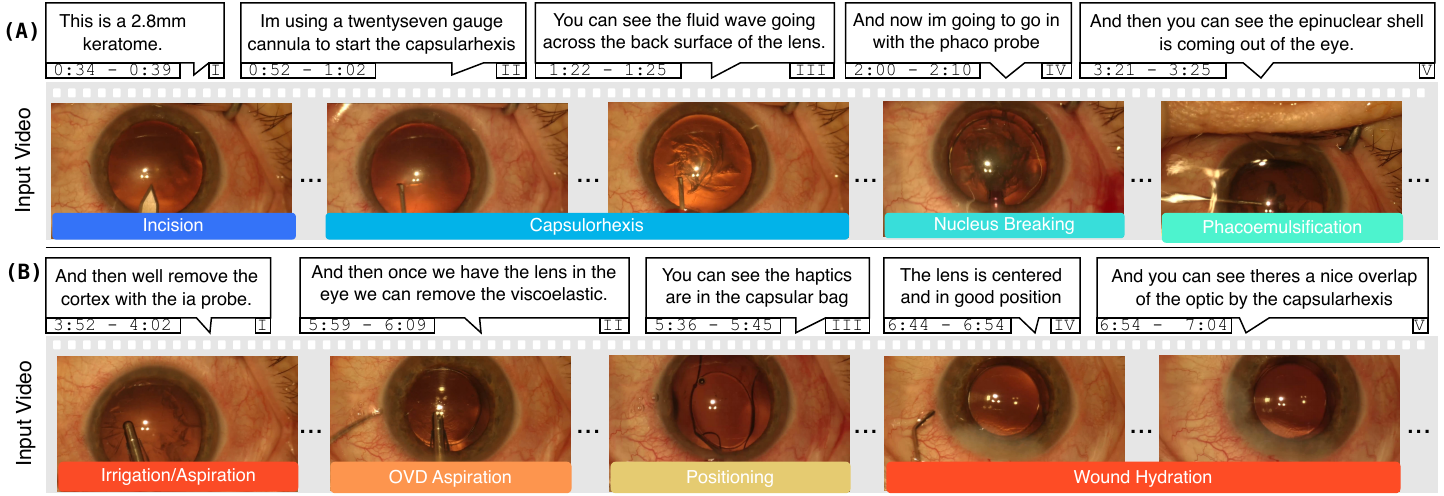}
    \caption{
    Example captions and phase predictions on video \textit{test\_08.mp4} of the CATARACTS test set. The fully captioned video is available in the supplementary materials.
    }
    \label{fig:caps_qual_08}
\end{figure}

Figures~\ref{fig:caps_qual_02} and \ref{fig:caps_qual_08} provide example captions generated by the model for videos \textit{test\_02.mp4} and \textit{test\_08.mp4} from the CAT test set, respectively. In \textit{test\_02.mp4}, the model shows strong performance, correctly identifying tools like the keratome (\texttt{(A)I}) and actions such as creating an incision. It also accurately describes fine details, such as bleeding at the incision site (\texttt{(A)II}), demonstrating a fine-grained understanding of the scene. However, some issues are noticeable, such as hallucinations about tool sizes, for example, the 27-gauge cannula in \texttt{(A)III}, which the model cannot accurately determine without a proper frame of reference. Additionally, \texttt{(A)IV} repeats a description unnecessarily, showing a common issue of repeated text. The model performs well when describing intricate steps, as seen in \texttt{(B)I}, and it even explains the causes behind later complications like iris prolapse (\texttt{(B)II}) before identifying the complication itself (\texttt{(B)III}).

In \textit{test\_08.mp4}, the model continues to provide detailed descriptions, accurately identifying tools and surgical conditions (\texttt{(A)III, IV}) but again hallucinating tool sizes (\texttt{(A)I, II}). It also reflects a dataset issue in \texttt{(B)I}, where inconsistent naming leads it to call the irrigation/aspiration probe an “ia probe,” a common abbreviation. Another limitation appears in \texttt{(B)IV}, where the model focuses on describing lens placement instead of the main action, wound hydration, likely due to biases in the training dataset commentary that prioritize lens positioning. Finally, the surgical outcome is summarized in \texttt{(B)V}.

While the model faces challenges such as hallucinating tool sizes, repeating descriptions, handling inconsistent terminology, and focusing on the right actions in some cases, it represents a significant advancement in long-term dense video captioning for the surgical domain. It demonstrates exceptional capabilities in identifying surgical tools, actions, and their relationships, generating clear and coherent captions across all procedural steps. Notably, it excels in detecting rare complications—a remarkable achievement given the absence of a dedicated DVC dataset for any surgical domain. These results lay a strong foundation for future advancements in surgical video analysis.

\end{document}